\documentclass[11pt,a4paper,table,xcdraw]{article}
\usepackage[hyperref]{acl2020}
\usepackage{times}
\usepackage{latexsym}
\usepackage{xcolor}

\usepackage{multirow}
\usepackage{color, colortbl}
\usepackage{xcolor}
 \usepackage{times}

\usepackage{xcolor}

\usepackage{booktabs}
\usepackage{setspace}

\usepackage{microtype}

\aclfinalcopy 

\title{Gender in Danger?\\ {E}valuating Speech Translation Technology on the {M}u{ST-SHE} Corpus}

\author{Luisa Bentivogli$^1$\thanks{  \textcolor{white}{$\ast$}These authors contributed equally. The work by Beatrice Savoldi was carried out
during an internship at Fondazione Bruno Kessler.} , Beatrice Savoldi$^2$\footnotemark[1] , Matteo Negri$^1$, \\ \textbf{Mattia Antonino Di Gangi}$^{1,2}$\textbf{, Roldano Cattoni}$^1$\textbf{, Marco Turchi}$^1$ \\
  $^1$Fondazione Bruno Kessler \\
  $^2$University of Trento \\
   Trento, Italy \\
  {\tt \{bentivo,negri,digangi,cattoni,turchi\}@fbk.eu} \\
\texttt{beatrice.savoldi@unitn.it} \\}

\date{}

\begin{document}
\maketitle
\begin{abstract}
Translating from languages without productive grammatical gender like English into gender-marked languages is  a well-known difficulty for machines. This difficulty is also due to the fact that the training data on which models are built typically reflect the asymmetries of natural languages, gender bias included. Exclusively fed with \textit{textual data}, machine translation is intrinsically constrained by the fact that the input sentence does not always contain clues about the gender identity of the referred human entities.  But what happens with speech translation, where the input is an \textit{audio} signal? Can audio provide additional information to reduce gender bias? We present the first thorough investigation of gender bias in speech translation, contributing with: \textit{i)} the release of a benchmark useful for future studies, and \textit{ii)} the comparison of different technologies (cascade and end-to-end) on two language directions (English-Italian/French).
\end{abstract}

\section{Introduction}
\label{sec:intro}
With the exponential popularity of deep learning approaches for a great range of natural language processing (NLP) tasks being integrated in our daily life, the need to address the issues of  \textit{gender fairness}\footnote{We  acknowledge that \textit{gender} is a multifaceted notion, not necessarily constrained within binary assumptions. However, since speech translation is hindered by the scarcity of available data, we rely on the female/male distinction of gender, as it is linguistically reflected in existing natural data.} and \textit{gender bias} has become a growing interdisciplinary concern. Present-day studies on a variety of NLP-related  tasks,  such as sentiment analysis \citep{kiritchenko-mohammad-2018-examining} coreference resolution \citep{rudinger-etal-2018-gender,webster-etal-2018-mind,zhao-etal-2018-gender}, visual semantic-role labeling \citep{zhao-etal-2017-men} or language modeling \citep{lu2019gender}, attest the existence of a systemic bias that reproduces gender stereotypes discriminating women. In translation-related tasks, gender bias  arises from the extent through which each language formally expresses the female or male gender of a referred human entity. Languages with a grammatical system of gender, such as Romance languages,  rely on a copious set of morphological (inflection) and syntactic (gender agreement) devices applying to numerous parts of speech \citep{Hockett:58}. Differently, English is a natural gender language that only reflects distinction of sex via pronouns, inherently gendered  words (\textit{boy, girl}) and exceptionally with marked nouns (\textit{actor, actress}). For all the other indistinct neutral words, the gender of the referred entity -- if available -- is inferred from   contextual information present in the discourse, e.g. \underline{he}/\underline{she} is a \textit{friend}. 

Nascent inquiries on machine translation (MT) pointed out that machines tend to reproduce the linguistic asymmetries present in the real-world data  they are trained on. In the case of gender inequality, this is made apparent by  the attribution of occupational roles from gender-neutral linguistic forms into marked ones, where MT often wrongly chooses male-denoting (pro)nouns, e.g. identifying \textit{scientist}, \textit{engineer} or \textit{doctor} as men \citep{Prates2018AssessingGB,escude-font-costa-jussa-2019-equalizing}. Failing to pick the appropriate feminine form is both a technical and an ethical matter:  gender-related errors affect the accuracy of MT  systems but, more significantly, a biased system can dangerously perpetuate the under-/misrepresentation of a demographic group \citep{crawford2017trouble}.

Previous studies accounting for MT systems' strengths and weaknesses in the translation of gender shed light on the problem but, at the same time, have limitations. On one hand, the existing evaluations focused on gender bias were largely conducted on challenge datasets, which are controlled artificial benchmarks that provide a limited perspective on the extent of the phenomenon and may force unreliable conclusions  \citep{Prates2018AssessingGB,cho-etal-2019-measuring,escude-font-costa-jussa-2019-equalizing,stanovsky-etal-2019-evaluating}.  On the other hand, the natural corpora built on conversational language that were used in few studies \citep{Elaraby2018GenderAS,vanmassenhove-etal-2018-getting} include only a restricted quantity of not isolated gender-expressing forms, thus not permitting either extensive or targeted evaluations. Moreover, no attempt has yet been made to assess if and how \textit{speech} translation (ST) systems are affected by this particular problem. As such, whether ST technologies that leverage audio inputs can retrieve useful clues for translating gender in addition to contextual information present in the discourse, or supply for their lack, remains a largely unexplored question.  In the light of above, the contributions of this paper are:

\textbf{(1)} 
We present the first systematic analysis aimed to assess ST performance on gender translation. To this aim, we compare the state-of-the-art cascaded approach with the emerging end-to-end paradigm, investigating their ability to properly handle different categories of gender phenomena.

\textbf{(2)} 
We publicly release MuST-SHE,\footnote{MuST-SHE is released under a CC BY NC ND 4.0 International license, and is freely downloadable at \url{ict.fbk.eu/must-she}.} a multilingual, natural benchmark allowing for a fine-grained analysis of gender bias in MT and ST. MuST-SHE is a subset of the TED-based MuST-C corpus \citep{di-gangi-etal-2019-must} and is available for English-French and English-Italian.\footnote{The current release of the corpus also includes an English-Spanish section, which was completed in April 2020.} For each language pair, it comprises  $\sim$1,000 (\textit{audio}, \textit{transcript}, \textit{translation}) triplets annotated with qualitatively  differentiated and balanced gender-related phenomena.

\textbf{(3)}  We implement a new evaluation  method that acknowledges and adapts previous related works to go beyond them and make BLEU scores informative about gender. It removes unrelated factors that may affect the overall performance of a system to soundly estimate gender bias. 

On the two language pairs addressed, our comparative evaluation of cascade \textit{vs.} end-to-end ST  systems indicates that the latter are able to better exploit audio information to  translate specific gender phenomena, for which the cascade systems require externally-injected information.

\section{Background}
\label{sec:relwork}
\textbf{Speech translation.}
The task of translating audio speech in one language into text in another language has been traditionally approached with cascade    architectures combining  automatic speech recognition (ASR) and MT components \citep{eck2005overview}.  The main advantage of this pipelined solution is that it can directly plug-in state-of-the-art technology for both components and exploit the wealth of training data available for the two tasks. The approach, however, has  some drawbacks. One is error propagation: sub-optimal transcriptions by the ASR component have significant impact on the final output produced by the MT component. To cope with this issue, recent works focused on making MT models more robust to noisy input transcripts \citep{sperber2017toward,sperber-etal-2019-self,digangi2019robust}.

A second issue,  particularly relevant to this research, is the information loss when passing from audio to text representations. Even with perfect transcripts, subtle aspects  that cannot be grasped from the text only (e.g. speaker's pitch as a clue of his/her gender) can only be reintroduced by injecting external knowledge to support  the MT step \citep{Elaraby2018GenderAS}. By avoiding intermediate text representations,  direct end-to-end  translation from audio to text \citep{berard:hal-01408086} can potentially cope with these limitations. However, due to the dearth of  training corpora, it still underperforms with respect to the cascaded approach. Recent evaluation campaigns  \citep{jan2018iwslt,jan2019iwslt}  have  shown that, although the gap is gradually closing (less than 2 BLEU points), cascade models  still represent the state-of-the-art. In spite of the steady technological progress,  little has so far been done to directly compare the two technologies on specific translation problems like the one  addressed in this paper.

\noindent \textbf{Measuring gender bias.}
Previous attempts to test the production of gender-aware automatic translations solely focused on MT,  where a  widespread approach  involves the creation of challenge datasets focused on specific linguistic phenomena. \citet{Prates2018AssessingGB} and \citet{cho-etal-2019-measuring} construct  template sentences using occupational or sentiment words associated with a gender-neutral pronoun, to be translated into an English gender-specified one (\textit{\textbf{{\textit[{{x}]}}} is a professor: \textbf{{\textit{he/she}}} is a professor}).  Similarly, the Occupations Test \citep{escude-font-costa-jussa-2019-equalizing} and Wino MT \citep{stanovsky-etal-2019-evaluating} cast human entities into proto- or anti-stereotypical gender associations via coreference linking (e.g. the English sentence \textit{``The janitor does not like \textbf{the baker} because \underline{she}/\underline{he} always messes up the kitchen''}, where \textit{``the baker''} is to be translated into Spanish as \textit{\textbf{la panadera}} or \textit{\textbf{el panadero}} depending on the English pronoun). Although such simple constructions allow for targeted experiments, artificial data characterized by a qualitatively limited variety of phenomena generate constrained environments that may produce biased results. As far as studies on naturally occurring data are concerned, \citet{vanmassenhove-etal-2018-getting} estimate MT systems' performance in the realization of speaker's gender agreement on two male and female test sets containing first person singular pronouns. This strategy increases the chances to isolate speaker-dependent gendered expressions, but still, the employed BLEU metric does not pointedly grasp the effect of gender translation on the output, as the overall performance is also impacted by other factors.  Analogously, \citet{Elaraby2018GenderAS}  design a set of agreement rules to automatically recover 300 gender-affected sentences in their corpus, but the evaluation relies on global BLEU scores computed on a bigger  set (1,300 sentences) and does not consider male-female related differences.  \citet{moryossef-etal-2019-filling} use a parser to detect morphological realizations of speakers' gender on a single female-speaker corpus that does not permit inter-gender comparisons. 

In light of above, an ideal test set should consist of naturally occurring data exhibiting a diversified assortment of gender phenomena so to avoid forced predictions with over-controlled procedures. Also, a consistent amount of equally distributed feminine and masculine gender realizations need to be identified to disentangle the accuracy of gender translation from the overall model's performance. Accordingly, in  $\S$\ref{sec:dataset} we present MuST-SHE, a multilingual test set designed for the  investigation of gender bias in ST, which, as explained in $\S$\ref{sec:exp}, is used for a targeted gender-sensitive evaluation approach. 

\section{The MuST-SHE benchmark}
\label{sec:dataset}
We built MuST-SHE on naturally occurring data retrieved from MuST-C \citep{di-gangi-etal-2019-must}, the largest freely available multilingual corpus for ST, which comprises (\textit{audio}, \textit{transcript}, \textit{translation}) triplets extracted from TED talks data. Besides being multilingual,  MuST-C is characterized by high-quality speech and a variety of different speakers that adequately represent women, two aspects that determined its selection among other existing corpora \citep{post2013improved,kocabi18, sanabria18how2}.  As such, MuST-SHE was compiled by targeting in the original dataset linguistic phenomena that entail a gender identification from English into Italian and French, two Romance languages that extensively express gender via feminine or masculine morphological markers on nouns, adjectives, verbs and other functional words (e.g. articles and demonstratives).

\subsection{Categorization of gender phenomena}
\label{ssec:data-annotations}
MuST-SHE is compiled with segments that require the translation of at least one English gender-neutral word into the corresponding masculine or feminine target word(s), where such formal distinction semantically conveys and conflates with an actual distinction of sex \citep{Corbett:91}. For instance, the English utterance ``\textit{a good teacher}" would either become in French ``\textit{un bon enseignant}'' or ``\textit{une bonne enseignante}'' for, respectively, a male or female referent. In spoken language data, the  human entity that determines gender agreement is either the speaker him/herself (\textit{I am a good teacher}) or another person the speaker is referring to  (\textit{he/she is a good teacher}).  We classify our phenomena of interest in two categories based on where the necessary information to disambiguate gender can be recovered, namely (\textbf{Category 1}) from the audio signal, when gender-agreement only depends on the speaker's gender, which can be captured from intrinsic properties of the audio (\textit{I am a teacher} uttered by a man/woman); (\textbf{Category 2}) from the utterance content, where contextual hints such as gender-exclusive words (\textit{mom}),  pronouns (\textit{she, his}) and proper nouns (\textit{Paul}) inform about the gender of the referent.

\subsection{Dataset creation and annotation}
\label{ssec:data-creation}

\begin{table*}[h!]
\centering
\footnotesize
\begin{tabular}{lllc}
  \toprule
 \textbf{Form}
 &  & \textbf{Category 1}: \textit{Gender info in audio} & \textbf{Speaker}\\ [1.mm]
 \cmidrule{2-3}
Fem. & \textsc{src} & I was \textbf{born} and \textbf{brought up }in Mumbai. & Female \\
& \textsc{C-Ref}$_{It}$  &  Sono \textbf{nata} e \textbf{cresciuta} a Mumbai. &\\ 
& \textsc{W-Ref}$_{It}$ & Sono \textbf{nato} e \textbf{cresciuto} a Mumbai.&\\ [1.5mm]
& \textsc{src} & I was \textbf{born} and brought up in Mumbai. &  \\
& \textsc{C-Ref}$_{Fr}$ &  Je suis \textbf{n\'ee} et j'ai  grandi \`a Mumbai. &\\ 
& \textsc{W-Ref}$_{Fr}$ & Je suis \textbf{n\'e} et j'ai  grandi \`a Mumbai. &\\ [1.5mm]
 \cmidrule{2-3}
Masc. & \textsc{src} & I \textbf{myself} was \textbf{one} of them, and this is what I talk about at the HALT events. & Male\\
& \textsc{C-Ref}$_{It}$ & Io \textbf{stesso} ero \textbf{uno} di loro, e parlo di questo agli eventi HALT. &\\
& \textsc{W-Ref}$_{It}$ & Io \textbf{stessa} ero \textbf{una} di loro, e parlo di questo agli eventi HALT.  &\\ [1.5mm]
& \textsc{src} & I myself was \textbf{one} of them, and this is what I talk about at the HALT events. &\\
& \textsc{C-Ref}$_{Fr}$ & Moi-m\^{e}me, j'ai \'et\'e l'\textbf{un} d'eux, et voil\`a de quoi je parle aux \'ev\'enements d'HALT. &\\
& \textsc{W-Ref}$_{Fr}$ & Moi-m\^{e}me, j'ai \'et\'e l'\textbf{une} d'eux, et voil\`a de quoi je parle aux \'ev\'enements d'HALT. &\\ [1.5mm]

 \midrule
  & &  \textbf{Category 2}: \textit{Gender info in 
utterance content} &\\ [1.mm]
   \cmidrule{2-3}
   Fem. & \textsc{src} &    She'd get together with two  \textbf{of her dearest friends}, \textbf{these older} \underline{women}... & Male\\ 
   & \textsc{C-Ref}$_{It}$ & Tornava per incontrare un paio \textbf{delle sue} pi\`u \textbf{care amiche}, \textbf{queste} signore \textbf{anziane}...\\
   & \textsc{W-Ref}$_{It}$ & Tornava per incontrare un paio \textbf{dei suoi} pi\`u \textbf{cari amici}, \textbf{questi} signore \textbf{anziani}...\\ [1.5mm]
   & \textsc{src} &    She'd get together with two of her \textbf{dearest friends}, these \textbf{older} \underline{women}... & \\ 
& \textsc{C-Ref}$_{Fr}$ &  Elle se r\'eunissait avec deux de ses \textbf{amies} les plus \textbf{ch\`eres}, ces femmes plus \textbf{\^{a}g\'ees}... & \\
 & \textsc{W-Ref}$_{Fr}$ &  Elle se r\'eunissait avec deux de ses \textbf{amis} les plus \textbf{chers}, ces femmes plus \textbf{\^{a}g\'es}... & \\ [1.5mm]
\cmidrule{2-3}
Masc. & \textsc{src} &  \underline{Dean}  Kamen, \textbf{one of the}  great DIY \textbf{innovators}. \underline{His} technology...&  Female \\ 
& \textsc{C-Ref}$_{It}$ & Dean Kamen, \textbf{uno dei} più grandi \textbf{innovatori} del fai-da-te. La sua tecnologia… \\
& \textsc{W-Ref}$_{It}$ & Dean Kamen, \textbf{una delle} più grandi \textbf{innovatrici} del fai-da-te. La sua tecnologia…\\ [1.5mm]
& \textsc{src} &  \underline{Dean}  Kamen, \textbf{one} of the \textbf{great} DIY \textbf{innovators}. \underline{His} technology...&   \\
& \textsc{C-Ref}$_{Fr}$ &  Dean Kamen, l'\textbf{un} des \textbf{grands innovateurs} autonomes. Sa technologie... & \\
& \textsc{W-Ref}$_{Fr}$ &  Dean Kamen, l'\textbf{une} des \textbf{grandes innovatrices} autonomes. Sa technologie... & \\ [1.5mm]
\midrule
\end{tabular}
\caption{MuST-SHE annotated segments organized per category. For each example in En-It and En-Fr, the Correct Reference Translation (C-REF) shows the realization of target gender-marked forms (Masc/Fem) corresponding to English gender-neutral words in the source (SRC). In the Wrong Reference Translation (W-REF), Italian and French gender-marked words are swapped to their opposite gender form. The last column of the table provides information about the speaker's gender (Male/Female).}
\label{tab:Examples}
\end{table*}

To gain a better insight into MuST-C linguistic data and capture the features of gender, we initially conducted a qualitative cross-lingual analysis on 2,500 parallel sentences randomly sampled from the corpus. The analysis led to the  design of an automatic approach aimed  to  quantitatively and qualitatively  maximize the extraction of an assorted variety of gender-marked phenomena belonging to categories 1 and 2. Regular expressions were employed to transform gender-agreement rules into search patterns to be applied to MuST-C. Our queries were designed and adapted to the targeted language  pairs, categories, and masculine/feminine forms. To specifically match a differentiated range of gender-marked lexical items, we also compiled two series of 50 human-referring adjectives in French and Italian, as well as a list with more than 1,000 English occupation nouns obtained from the US Department of Labour Statistics\footnote{\url{http://www.bls.gov/emp/tables/emp-by-detailed-occupation.htm}} \citep{Prates2018AssessingGB}.

For each language direction, the pool of sentence pairs retrieved from MuST-C  was manually checked in order to: \textit{i)} remove noise and keep only pairs containing at least one gender phenomenon,  \textit{ii)} include all En-It/En-Fr corresponding pairs to create a common multilingual subset, and \textit{iii)} select the remaining pairs ensuring a balanced distribution of categories, feminine/masculine forms, and female/male speakers. Once the textual part of MuST-SHE was created, all the corresponding audio segments were manually checked in order to correct possible misalignments.

The resulting dataset was then manually enriched with different types of information that allow for fine-grained evaluations.  Annotations include: category, masculine/feminine form, speaker's gender, and all the  gender-marked expressions in the reference translation. Finally, in order to perform a sound evaluation able to discriminate  gender-related issues from other non-related factors that may affect  systems' performance, for each correct reference translation (\textsc{C-Ref}) we created an almost identical ``wrong'' alternative (\textsc{W-Ref}) in which all the gender-marked words are swapped to their opposite form (details in $\S$\ref{sec:exp}). Some examples extracted from MuST-SHE  are presented in Table \ref{tab:Examples}.

To ensure data quality, the whole dataset was created and annotated by an expert linguist with a background in translation studies, who  produced strict and comprehensive guidelines based on the preliminary manual analysis of a sample of MuST-C data (2,500 segments). Then, a second linguist independently re-annotated each MuST-SHE  segment with the corresponding category and produced an additional ``wrong'' reference. Being the annotation per category a straightforward task, it resulted in no disagreement for Category 1 and around 0.03\% for Category 2. Such few cases were removed from the dataset, which thus contains only segments in complete agreement. Disagreements were more common in the ``wrong'' references, since the task requires producing subtle variations that can be hard to spot. Disagreements, amounting to around 11\%, were all oversights and thus reconciled.

\subsection{Dataset statistics}
\label{ssec:data-stats}
MuST-SHE comprises 2,136 (\textit{audio}, \textit{transcript}, \textit{translation}) triplets (1,062 for En-It and 1,074 for En-Fr) uttered by 273 different speakers. A common subset of 696 instances allows for comparative evaluations across the two language directions. As shown by the statistics in Table \ref{tab:stats}, the corpus presents a balanced distribution across \textit{i)} masculine and feminine forms, and  \textit{ii)} gender phenomena per category. Female and male speakers  (558/513 for En-It, 577/498 for En-Fr) are  substantially balanced. The gender of the speaker and of the referred  entity in the utterance is the same in Category 1 (where the speakers talk about themselves),  while it differs in about 50\% of the segments in Category 2  (where they refer to other entities). 

MuST-SHE differs from standard test sets, as it is precisely designed to:  \textit{i)}  equally distribute gender references as well as speakers, and \textit{ii)} allow for a sound and focused evaluation on the accuracy of gender translation. As such, it satisfies the parameters to be qualified as a GBET, \textit{Gender Bias Evaluation Testset} \citep{sun-etal-2019-mitigating}, and represents the very first of its kind for  ST and MT created on natural data. 

\begin{table}[h!]
    \centering
    \begin{small}
    \begin{tabular}{l|ccc||ccc}
    \hline
      &  \multicolumn{3}{c||}{\hspace*{0.2em}\textbf{En-It}\hspace*{0.2em}} & \multicolumn{3}{c}{\hspace*{0.2em}\textbf{En-Fr}\hspace*{0.2em}} \\[.5mm]
        & Fem & Masc & \textit{Tot.} & Fem & Masc  & \textit{Tot.}\\ [1.mm]
    \hline 
    \hline
   Cat. \emph{1} & 278 & 282 & \textit{560} & 316 & 296 & \textit{612}\\  [1.mm]
    Cat.  \emph{2} & 238 & 264 & \textit{502} & 226 & 236 & \textit{462}\\  [1.mm]
    \cline{2-3}  \cline{5-6}
     \textit{Tot.} & \textit{516} & \textit{546} &   & \textit{542} & \textit{532} & \\ [1.mm]
     \hline  
     \textbf{Total} & \multicolumn{3}{c||}{\textbf{1,062} (1,940)} & \multicolumn{3}{c}{\textbf{1,074}  (2,010)} \\
    \hline
\end{tabular}
\end{small}
    \caption{MuST-SHE statistics. En-It and En-Fr number of segments split into feminine and masculine gender phenomena and category. In parentheses the total number of  gender-marked words annotated in the references.}
    \label{tab:stats}
\end{table}

\section{Experimental Setting}
\label{sec:exp}

\subsection{Evaluation Method}
\label{ssec:method}
MT evaluation metrics like BLEU \citep{papineni2002bleu} or TER \citep{Snover:06} provide a global score about translation ``quality'' as a whole. Used as-is, their holistic nature hinders the precise evaluation of systems' performance on an individual phenomenon as gender translation, since the variations of BLEU score are only a coarse and indirect indicator of better/worse overall performance \citep{callison-burch-etal-2006-evaluation}. This represents a limitation of recent related works, which over-rely on the results of a BLEU-based quantitative analysis. For instance, the BLEU  gains obtained by prepending gender tags or other artificial  antecedents to the input source, as in \citet{vanmassenhove-etal-2018-getting} and \citet{moryossef-etal-2019-filling}, cannot be assuredly ascribed to a better control of gender features. To overcome this problem, \citet{moryossef-etal-2019-filling} complement their discussion with a qualitative syntactic   analysis, which implies the availability of a parser for the target language and a higher complexity of the whole evaluation protocol. Instead, our aim is to  keep using BLEU\footnote{Still the \textit{de facto} standard in MT evaluation in spite of constant research efforts towards metrics that better correlate with human judgements.}  and make the resulting scores informative about systems' ability to produce the correct gender forms.

To this aim, for each reference $c$ in the corpus we create a ``wrong'' one  that is identical to $c$, except for the morphological signals that convey gender agreement.  In particular, for each gender-neutral English word in the source utterance  (e.g. \textit{``one''}, \textit{``great''} and \textit{``innovators''} in the $4^{th}$  example of Table \ref{tab:Examples}), the correct  translation (containing the  French words with masculine inflection \textit{``un''}, \textit{``grands''} and \textit{``innovateurs''})  is swapped into its opposite gender form (containing feminine-marked words \textit{``une''}, \textit{``grandes''} and \textit{``innovatrices''}). The result is a new set of references that, compared to the correct ones, are ``wrong'' only with respect to the formal expression of gender.

The underlying idea is that, as the two reference sets differ only for the swapped  gendered forms, results' differences for the same set of hypotheses produced by a given system can  measure its capability to handle gender phenomena. In particular, we argue that higher values on the wrong set can  signal a potentially gender-biased  behaviour.  In all the cases where the required gender realization is \textit{feminine}, significantly higher BLEU results computed on the wrong set would signal a bias towards producing masculine forms, and vice versa.  Although this idea recalls the gender-swapping approach used in previous NLP studies on gender bias \citep{sun-etal-2019-mitigating,lu2019gender,kiritchenko-mohammad-2018-examining,zhao-etal-2018-gender,cao2019genderinclusive}, in such works it is only applied  to pronouns; here we extend it to any gender-marked part of speech.

In addition to the quantitative  BLEU-based evaluation\footnote{We also computed TER scores and the results are fully in line with the reported BLEU scores.}, we also perform a  fine-grained  qualitative analysis of systems' accuracy in producing the target gender-marked words. We compute accuracy as the proportion of gender-marked words in the references that are correctly translated by the system. An upper bound of one match for each gender-marked word is applied in order not to reward over-generated terms. Besides global accuracy, we also compute scores on both the correct and the wrong reference sets, as well as  per category. 

It's worth remarking that the BLEU-based and the accuracy-based evaluations are complementary: the former aims to shed light on system's translation performance with respect to gender phenomena; the latter, which  is more discriminative, aims to point to the actual words through which gender is realized. Compared to the standard BLEU-based evaluation with correct references only, we expect that the possible differences suggested by its extension with gender swapping will be reflected and amplified by sharper accuracy differences.

\subsection{ST Systems}
\label{ssec:systems}
In our experiments, we  compare an \texttt{End2End} system with two cascade systems (\texttt{Cascade} and \texttt{Cascade+tag}), whose architectures are described below.

Our \texttt{\textbf{End2End}} system uses the S-transformer architecture, which has proved to work reasonably well for this task \citep{digangi2019adapting}. It is an encoder-decoder architecture that modifies the Transformer architecture~\citep{vaswani2017attention} in two aspects.  First, the audio input -- in the form of sequences of 40 MFCCs~\citep{davis1980comparison}  -- is processed by a stack of 2D CNNs \citep{lecun1998gradient}, each followed by batch normalization \citep{ioffe2015batch} and ReLU  nonlinearity. Second,  the output of the CNNs is processed by 2D self-attention networks to provide a larger context to each element. The output of the 2D attention is then summed with the positional encoding and fed to transformer encoder layers. In the second part, a distance penalty is added to the non-normalized probabilities in the encoder self-attention networks in order to bias the computation towards the local context. To improve translation quality, the \texttt{End2End} systems are trained on the MuST-C and Librispeech  \citep{kocabi18} corpora using SpecAugment \citep{park2019specaugment}. Since Librispeech is a corpus for ASR, we augmented it  by automatically  translating the original English transcripts into both target languages.  Translations are performed at character level, using the MT systems integrated in the cascade model.

\begin{table*}[h!]
  \centering
  \small
  \begin{tabular}{l|l||ccc||ccc|ccc}
    \hline
    &Systems &  \multicolumn{3}{c||}{\hspace*{0.2em}\textbf{All}\hspace*{0.2em}} & \multicolumn{3}{c|}{\hspace*{0.2em}\textbf{Feminine}\hspace*{0.2em}} & \multicolumn{3}{c}{\hspace*{0.2em}\textbf{Masculine}\hspace*{0.2em}} \\
      &    &  Correct & Wrong &  Diff & Correct & Wrong &  Diff & Correct & Wrong & Diff \\
    \hline\hline
  \multirow{3}{*}{En-It} & \texttt{End2End}  &  21.5  & 19.7 &   1.8\cellcolor[HTML]{EFEFEF} &   20.2    &  19.3  &  0.9 \cellcolor[HTML]{EFEFEF} &  22.7 & 20.0  &  2.7\cellcolor[HTML]{EFEFEF} \\
    \cline{2-11} 
   & \texttt{Cascade}     & 24.1 & 22.4 & 1.8\cellcolor[HTML]{EFEFEF} &  22.8  & 21.9 &  0.8 \cellcolor[HTML]{EFEFEF} & 25.5 & 22.8 &  2.7\cellcolor[HTML]{EFEFEF}\\
  &  \texttt{Cascade+Tag}     & 23.8 & 20.9 & 2.9\cellcolor[HTML]{EFEFEF} &  23.0  & 20.4 &  2.6\cellcolor[HTML]{EFEFEF} & 24.5 & 21.3 &  3.2\cellcolor[HTML]{EFEFEF}\\
    \hline
    \hline\hline
   \multirow{3}{*}{En-Fr}  & \texttt{End2End}   &  27.9 & 25.8 &  2.1\cellcolor[HTML]{EFEFEF} & 26.3 & 25.0 &  1.3\cellcolor[HTML]{EFEFEF} & 29.5 & 26.4 &  3.1\cellcolor[HTML]{EFEFEF}\\ 
    \cline{2-11} 
 &   \texttt{Cascade}     & 32.2 & 30.1  &  2.1\cellcolor[HTML]{EFEFEF} & 30.4  & 29.4 &  1.0\cellcolor[HTML]{EFEFEF} & 33.8 & 30.8 &  3.0\cellcolor[HTML]{EFEFEF}\\
  &  \texttt{Cascade+Tag}   &  32.2 & 28.6  &  3.6\cellcolor[HTML]{EFEFEF} & 31.6   &  28.0 &  3.6\cellcolor[HTML]{EFEFEF} & 32.7  & 29.2 & 3.5\cellcolor[HTML]{EFEFEF} \\
    \hline
  \end{tabular}
  \caption{BLEU scores for En-It and En-Fr on MuST-SHE. Results are provided for the whole dataset (All) as well as split according to feminine and masculine word forms. Results are calculated for both the \textit{Correct} and \textit{Wrong} datasets, and their difference is provided (Diff).}
  \label{tab:bleu}
 \end{table*}

Our \texttt{\textbf{Cascade}} systems share the same core (ASR, MT) technology.  The ASR component is based on the KALDI toolkit \citep{povey2011kaldi}, featuring a time-delay neural network and lattice-free maximum mutual information discriminative sequence-training ~\citep{povey2016}. The audio data for acoustic modeling include the clean portion of LibriSpeech \citep{librispeech} ($\sim$460h) and a variable subset of the MuST-C training set ($\sim$450h), from which 40 MFCCs per time frame were extracted; a MaxEnt  language model~\citep{Alumae2010} is estimated from the corresponding transcripts ($\sim$7M words). The MT component is based on the Transformer architecture, with parameters similar to those used in the original paper. The training data are collected from the OPUS repository,\footnote{\url{http://opus.nlpl.eu}} resulting in 70M pairs for En-It and 120M for En-Fr. For each language pair, the MT system is first trained on the OPUS data and then fine-tuned on MuST-C training data ($\sim$250K pairs) -- from which the MuST-SHE segments are removed. Byte pair encoding (BPE) \citep{sennrich2015neural} is applied to obtain 50K sub-word units. To mitigate error propagation and make the MT system more robust to ASR errors, similarly to \citep{digangi2019robust} we tune it on a dataset derived from MuST-C, which includes both human and automatic transcripts. The training set, consisting of (\textit{audio}, \textit{transcript}) pairs, is split in two equally-sized parts: the first one is used to adapt the ASR system to the TED talk language, while the second part is transcribed by the tuned ASR system. The human transcripts of the first half and the automatic transcripts of the second half  are concatenated and used together with their reference translations to fine-tune the MT system. This process makes the MT system aware of possible ASR errors and results in more than 2 BLEU points improvement on the MuST-C test set.

We also train an enhanced version  of the \texttt{Cascade} system. Similarly to \citet{vanmassenhove-etal-2018-getting}, it is informed about speaker's gender by pre-pending a gender token ($<$toM$>$ or $<$toF$>$) to each source transcript. The gender token is obtained by manually assigning the correct gender label to each speaker in MuST-C. This externally-injected knowledge allows the \texttt{Cascade+Tag} system to mimic end-to-end technology by  leveraging gender information during translation.

To check the overall quality of our systems, we compared them  with published results on  MuST-C test data.  Our \texttt{End2End} systems (En-It: 21.5, En-Fr: 31.0) outperform all the models proposed in \citet{digangi2019adapting}, which were trained only on MuST-C (En-It: end2end 16.8, cascade 18.9; En-Fr: end2end 26.9, cascade 27.9). Our \texttt{Cascade} (En-It: 27.4 En-Fr: 35.5) also outperforms the system described in \citet{indurthi2019data} (En-Fr: 33.7). Our results are in line with the findings of IWSLT 2019 \citep{jan2019iwslt} showing that the cascade approach still outperforms the direct one, although with a gradually closing gap.

\section{Results and Discussion}
\label{sec:results}
\textbf{BLEU}. Table \ref{tab:bleu} presents translation results in terms of BLEU score on the MuST-SHE dataset. Looking at overall translation quality (\textit{All/Correct} column), the results on both language pairs  show that the highest performance is achieved by cascade architectures, which are better than  \texttt{End2End} by 2.6 points for En-It and 4.3 for En-Fr. We do not observe a statistically significant difference between \texttt{Cascade} and \texttt{Cascade+Tag}, suggesting that the injection of gender information into   \texttt{Cascade+Tag}  does not have visible effects  in terms of  translation quality, even on a focused dataset like MuST-SHE where each segment contains at least one gender realization.  Our results thus seem to be in contrast with  previous works implementing the same injection approach \citep{vanmassenhove-etal-2018-getting,Elaraby2018GenderAS}.

However, looking at the scores' gap between the \textit{Correct} and the \textit{Wrong} datasets (\textit{All/Diff} column), it becomes evident that the standard evaluation based on BLEU calculated on a single correct reference hides specific relevant aspects in translation. In fact, despite the lower overall BLEU scores, for both language pairs \texttt{End2End} performs on par with \texttt{Cascade} as far as gender phenomena are concerned (1.8 on En-It and 2.1 on En-Fr). Also, the largest \textit{All/Diff} value achieved by the enhanced \texttt{Cascade+Tag} supports the results obtained in previous studies \citep{vanmassenhove-etal-2018-getting,Elaraby2018GenderAS}, confirming the importance of applying gender-swapping in BLEU-based evaluations focused on gender translation.

\begin{table*}[h!]
  \centering
  \small
  \begin{tabular}{l|l||ccc||ccc|ccc}
           \hline
    &Systems &  \multicolumn{3}{c||}{\hspace*{0.2em}\textbf{All}\hspace*{0.2em}} & \multicolumn{3}{c|}{\hspace*{0.2em}\textbf{Feminine}\hspace*{0.2em}} & \multicolumn{3}{c}{\hspace*{0.2em}\textbf{Masculine}\hspace*{0.2em}} \\
      &    &  Correct & Wrong & Diff & Correct & Wrong & Diff & Correct & Wrong & Diff \\
    \hline\hline
    \multirow{3}{*}{En-It} & \texttt{End2End}   &     43.3  &16.4  &  26.9\cellcolor[HTML]{EFEFEF} & 34.2    & 24.0   &  10.2\cellcolor[HTML]{EFEFEF}  & 51.3 & 9.6 & 41.7 \cellcolor[HTML]{EFEFEF}  \\
    \cline{2-11}
   & \texttt{Cascade}      & 41.1 & 17.5 & 23.6\cellcolor[HTML]{EFEFEF} &  33.7 & 24.5 & 9.2 \cellcolor[HTML]{EFEFEF} & 47.6 & 11.2 & 36.4 \cellcolor[HTML]{EFEFEF} \\
    & \texttt{Cascade+Tag}       & 48.0 & 10.4 &  37.6\cellcolor[HTML]{EFEFEF} &  44.7 & 14.0 & 30.7 \cellcolor[HTML]{EFEFEF}& 51.0 &  7.2 & 43.8 \cellcolor[HTML]{EFEFEF} \\
    \hline
        \hline\hline
     \multirow{3}{*}{En-Fr} & \texttt{End2End}   &   46.0 & 19.0 &  27.0\cellcolor[HTML]{EFEFEF} & 35.8 & 25.0 & 13.8\cellcolor[HTML]{EFEFEF}& 55.3 & 13.8 & 41.5\cellcolor[HTML]{EFEFEF}\\
    \cline{2-11}
   & \texttt{Cascade}      & 49.6 & 20.5 & 29.1 \cellcolor[HTML]{EFEFEF}&  39.6 & 26.2 & 13.4 \cellcolor[HTML]{EFEFEF}& 58.7  & 15.2 & 43.5 \cellcolor[HTML]{EFEFEF} \\
    & \texttt{Cascade+Tag}       & 57.2 & 11.3 & 45.9\cellcolor[HTML]{EFEFEF} &  53.8 & 11.8 & 42.0 \cellcolor[HTML]{EFEFEF} & 60.3 &  10.7  & 49.6 \cellcolor[HTML]{EFEFEF} \\ \hline
  \end{tabular}
  \caption{Accuracy scores for En-It and En-Fr on MuST-SHE. Results are provided for the whole dataset (All) as well as split according to feminine and masculine word forms. Results are calculated for both the \textit{Correct} and \textit{Wrong} datasets, and their difference is provided (Diff).}
  \label{tab:accuracy-all}
 \end{table*}

\begin{table*}[t]
\centering
\small
\begin{tabular}{|c|ccc|ccc| c |ccc|ccc|}
\cline{2-7} \cline{9-14}
 \multicolumn{1}{c|}{} & \multicolumn{6}{c|}{\textbf{En-it}} & \multicolumn{1}{c|}{\cellcolor[HTML]{FFFFFF}} & \multicolumn{6}{c|}{\textbf{En-Fr}} \\ \cline{2-7} \cline{9-14}
 \multicolumn{1}{c|}{} & \multicolumn{3}{c|}{\textbf{Feminine}} & \multicolumn{3}{c|}{\textbf{Masculine}} & 
 \multicolumn{1}{c|}{\cellcolor[HTML]{FFFFFF}} & \multicolumn{3}{c|}{\textbf{Feminine}} & \multicolumn{3}{c|}{\textbf{Masculine}} \\ \cline{2-7} \cline{9-14} 
 \multicolumn{1}{c|}{} & \multicolumn{6}{c|}{\cellcolor[HTML]{9B9B9B}{\color[HTML]{333333} \textbf{End2End}}} & \multicolumn{1}{c|}{\cellcolor[HTML]{FFFFFF}} & \multicolumn{6}{|c|}{\cellcolor[HTML]{9B9B9B}{\color[HTML]{333333} \textbf{End2End}}} \\ \cline{2-7} \cline{9-14}
 \multicolumn{1}{c|}{} & Corr. & Wrong & \cellcolor[HTML]{EFEFEF}Diff. & Corr. & Wrong & \cellcolor[HTML]{EFEFEF}Diff. &  
 \multicolumn{1}{c|}{\cellcolor[HTML]{FFFFFF}}
 & Corr. & Wrong & Diff. & Corr. & Wrong & \cellcolor[HTML]{EFEFEF}Diff. \\ \cline{1-7} \cline{9-14}

Cat. 1 &26.7  & 27.2 & -0.5 \cellcolor[HTML]{EFEFEF}& 46.3 & 6.8  & 39.5  \cellcolor[HTML]{EFEFEF}&
\multicolumn{1}{c|}{\cellcolor[HTML]{FFFFFF}} &
25.4  &  29.5 &  -4.1 \cellcolor[HTML]{EFEFEF} &  48.0 &  7.7  &  40.3 \cellcolor[HTML]{EFEFEF} \\
Cat. 2 & 40.6 & 20.5 & 20.1 \cellcolor[HTML]{EFEFEF} &53.9 & 10.9  & 43.0 \cellcolor[HTML]{EFEFEF}  &  
\multicolumn{1}{c|}{\cellcolor[HTML]{FFFFFF}} &
 45.0 &  20.3 &  24.7 \cellcolor[HTML]{EFEFEF} &  60.0 &  17.6 &  42.4 \cellcolor[HTML]{EFEFEF}\\ \cline{1-7} \cline{9-14}

\multicolumn{1}{c|}{} & \multicolumn{6}{c}{\cellcolor[HTML]{9B9B9B}\textbf{Cascade}} & \multicolumn{1}{c|}{\cellcolor[HTML]{FFFFFF}} & \multicolumn{6}{|c|}{\cellcolor[HTML]{9B9B9B}\textbf{Cascade}} \\ \cline{1-7} \cline{9-14}

Cat. 1 &	15.9   & 34.5 & -18.6 \cellcolor[HTML]{EFEFEF} &	40.0   & 12.0 & \multicolumn{1}{c|}{28.0 \cellcolor[HTML]{EFEFEF}} &
\multicolumn{1}{c|}{\cellcolor[HTML]{FFFFFF}} &
20.4  & 37.5 & -17.1\cellcolor[HTML]{EFEFEF} & 49.1 & 13.0 & 36.1\cellcolor[HTML]{EFEFEF} \\
Cat. 2 & 48.9	& 15.7 &  33.2 \cellcolor[HTML]{EFEFEF} &  51.2	& 10.8 & \multicolumn{1}{c|}{40.4 \cellcolor[HTML]{EFEFEF}} & 
\multicolumn{1}{c|}{\cellcolor[HTML]{FFFFFF}} &
56.3  & 15.6  & 40.7 \cellcolor[HTML]{EFEFEF} & 64.9 & 16.7 &48.2 \cellcolor[HTML]{EFEFEF} \\ \cline{1-7} \cline{9-14}

 \multicolumn{1}{c|}{} & \multicolumn{6}{c|}{\cellcolor[HTML]{9B9B9B}\textbf{Cascade+Tag}} & \multicolumn{1}{c|}{\cellcolor[HTML]{FFFFFF}} & \multicolumn{6}{|c|}{\cellcolor[HTML]{9B9B9B}\textbf{Cascade+Tag}} \\ \cline{1-7} \cline{9-14}

Cat. 1 &	43.0 &	10.4 &	32.6 \cellcolor[HTML]{EFEFEF} &	48.5	& 2.9	& \multicolumn{1}{c|}{45.6 \cellcolor[HTML]{EFEFEF}} &
\multicolumn{1}{c|}{\cellcolor[HTML]{FFFFFF}} &	
53.7 &7.0 & 46.7\cellcolor[HTML]{EFEFEF}  & 55.4 &  4.3 &  51.1 \cellcolor[HTML]{EFEFEF} \\
Cat. 2 & 46.9 &	15.5 &	31.4 \cellcolor[HTML]{EFEFEF} &	51.7	& 9.6	& \multicolumn{1}{c|}{42.1 \cellcolor[HTML]{EFEFEF}} &  
\multicolumn{1}{c|}{\cellcolor[HTML]{FFFFFF}} &
54.2 & 14.8 & 39.4 \cellcolor[HTML]{EFEFEF}& 62.8 & 15.5  & 47.3 \cellcolor[HTML]{EFEFEF} \\ \cline{1-7} \cline{9-14}
\end{tabular}
\caption{Accuracy scores for En-It and En-Fr split according to MuST-SHE categories (Cat 1: information in audio, Cat 2: information in utterance content). For each category, results are further split into masculine/feminine forms. Results are calculated for both the \textit{Correct} and \textit{Wrong} datasets, and their difference is provided (Diff).}
  \label{tab:accuracy-category}
\end{table*}

The fact that the \textit{All/Diff} values are always positive indicates that all the systems perform better on the \textit{Correct} dataset (i.e. they generate the correct gender-marked words more often than the wrong ones). However, examining the results at the level of masculine/feminine word forms, we notice that \textit{Diff} values are higher on the \textit{Masculine} subset (where the required gender realization is masculine) than in the \textit{Feminine} one (where the required gender realization is feminine). As discussed in $\S$\ref{ssec:method}, this signals a bias of the systems towards producing masculine forms.  The only exception is the En-Fr \texttt{Cascade+Tag}, where the \textit{Diff} values remain stable across the two subsets (3.6 and 3.5). This absence of bias towards the masculine forms is in line with the \textit{All/Diff} results indicating that this system is the best one in translating gender.

Although our gender-swapping methodology allows us to measure differences across systems that cannot be observed with standard BLEU evaluations, the results obtained so far may still conceal further interesting differences. This can depend on the fact that BLEU works at the corpus level and the small proportion of gender-marked words in MuST-SHE ($\sim$2,000 out of $\sim$ 30,000 total words, avg. 1.8 per sentence) can have limited influence on global measurements. To dig into these aspects,  our final analysis relies on accuracy, which is exclusively focused on gender-marked words. 

\noindent \textbf{Accuracy}. The results shown in Table \ref{tab:accuracy-all} are not only consistent with the BLEU ones, but also highlight differences that were previously indistinguishable. While the \textit{All/Diff} BLEU results for \texttt{End2End} and \texttt{Cascade} were identical on both languages, the \textit{All/Diff} accuracy scores show  that, although \texttt{End2End} performs better than \texttt{Cascade} for En-It, it performs worse for En-Fr. Also, with regards to \texttt{Cascade+Tag}, we can see that the \textit{Diff} value is higher on the \textit{Masculine} subset, thus showing that also this system is affected by gender bias, although to a lesser extent.

We now focus on   systems' results on the two categories represented in MuST-SHE: Category 1,  where the information necessary to disambiguate gender  can be recovered from the audio (speaker talking about him/herself) and Category 2,  where such information occurs in the utterance content (speaker talking about someone else). Results are shown in Table \ref{tab:accuracy-category}.
 
As for \textbf{Category 1}, \textit{Diff} values show that \texttt{Cascade} performance is the worst on both languages. This is due to the fact that its MT component cannot access the speaker's gender information necessary for a correct translation. This weakness becomes particularly evident in the \textit{Feminine} class, where the higher values on the \textit{Wrong} datasets (leading to negative values in columns \textit{Feminine/Diff}) highlight a 
strong
bias towards producing masculine forms.
Although still negative for the Feminine 
class, the   much better \textit{Diff} values obtained by  \texttt{End2End} show its ability to leverage audio features to correctly translate gender.  However, the gap with respect to \texttt{Cascade+Tag} -- by far the best system in Cat. 1 -- is still  large.  On one side, \texttt{End2End} might benefit from better audio representations. Indeed, as shown in~\citet{Kabil2018OnLT}, the MFCC features used by state-of-the-art models are not the most appropriate for gender recognition. On the other side, \texttt{Cascade+Tag} does not only take advantage of huge amounts of data to train its basic components, but it is also an oracle supported by the artificial injection of correct information about speakers' gender. 

In \textbf{Category 2}, where having direct access to the audio is not an advantage since gender information is present in the textual transcript, results show a different scenario. While  scores on the \textit{Masculine} class are not conclusive across languages, on the \textit{Feminine} class \texttt{End2End} always shows the worst performance. This can be explained by the fact that, being trained on a small fraction of the data used by the  cascade systems, \texttt{End2End} is intrinsically weaker and more prone to gender mistranslations. Also, it is noticeable that \texttt{Cascade+Tag} is slightly worse than \texttt{Cascade}, although the MT components are trained on the same amount of data. This is due to the dataset design choice (see $\S$\ref{ssec:data-stats}) to include $\sim$50\% of segments where the speaker's gender does not agree with the gender of the phenomenon to translate. This feature makes MuST-SHE particularly challenging for systems like \texttt{End2End} and \texttt{Cascade+Tag} since, in these specific cases, speaker's gender information (extracted from the source audio or artificially injected) is not relevant and can introduce noise.

All in all, translating gender is still an issue in ST and current technologies are affected by gender bias to variable extent. Through the analysis made possible by MuST-SHE, we have been able to  pinpoint their specific strengths and weaknesses  and pave the way for more informed future studies.

\section{Conclusion}
\label{sec:conclusion}
If, like human beings, ``machine learning is what it eats'', the different ``diet'' of  MT and ST models can help them to develop different skills. One is the proper treatment of gender, a problem when translating from languages without productive grammatical gender into gender-marked ones.  With respect to this problem, by eating parallel texts during training, MT performance is bounded by the statistical patterns learned from written material.  By eating (\textit{audio}, text) pairs, ST has a potential  advantage: the possibility to infer speakers' gender from  input audio signals. We  investigated for the first time the importance of this information in ST, analysing the behaviour of cascade (the state of the art in the field) and end-to-end ST technology (the emerging approach). To this aim, we created MuST-SHE, a benchmark annotated with different types of gender-related phenomena in two language directions.   Our evaluation shows that, in spite of lower overall performance,  the direct approach can actually exploit audio information to better handle speaker-dependent gender phenomena.  These are out of reach for cascade solutions, unless the MT step  is supplied with  external  (not always accessible) knowledge  about the speaker. Back to our title: if, in ST, gender is still in danger,  we encourage our  community to start its rescue from MuST-SHE and the findings discussed in this paper.

\section*{Acknowledgments}
This work is part of the project ``End-to-end Spoken Language Translation in Rich Data Conditions'',\footnote{\url{https://ict.fbk.eu/units-hlt-mt-e2eslt/}} which is financially supported by an Amazon AWS ML Grant. We thank our colleague Marco Matassoni for providing the the automatic transcripts used for our experiments with cascade systems.

\bibliography{FBK-acl2020,anthology}
\bibliographystyle{acl_natbib}




\end{document}